\documentclass{article}

\usepackage{arxiv}

\usepackage[utf8]{inputenc} 
\usepackage[T1]{fontenc}    
\usepackage{hyperref}       
\usepackage{url}            
\usepackage{booktabs}       
\usepackage{amsfonts}       
\usepackage{nicefrac}       
\usepackage{microtype}      
\usepackage{lipsum}
\usepackage{graphicx}
\usepackage[numbers]{natbib}
\usepackage{makecell}
\usepackage{amsmath}

\title{Data Augmentation and Resolution Enhancement using GANs and Diffusion Models for Tree Segmentation}

\author{
 Alessandro dos Santos Ferreira\thanks{Corresponding author: alessandro.ferreira@ufms.br} \\
  Federal University of Mato Grosso do Sul \\
  Campo Grande, MS, Brazil \\
   \And
 Ana Paula Marques Ramos \\
  São Paulo State University \\
  Presidente Prudente, SP, Brazil \\
  \And
 José Marcato Junior \\
  Federal University of Mato Grosso do Sul \\
  Campo Grande, MS, Brazil \\
 \And
 Wesley Nunes Gon\c calves \\
  Federal University of Mato Grosso do Sul \\
  Campo Grande, MS, Brazil \\
}

\begin{document}
\maketitle
\begin{abstract}
Urban forests play a key role in enhancing environmental quality and supporting biodiversity in cities. Mapping and monitoring these green spaces are crucial for urban planning and conservation, yet accurately detecting trees is challenging due to complex landscapes and the variability in image resolution caused by different satellite sensors or UAV flight altitudes.
While deep learning architectures have shown promise in addressing these challenges, their effectiveness remains strongly dependent on the availability of large and manually labeled datasets, which are often expensive and difficult to obtain in sufficient quantity.
In this work, we propose a novel pipeline that integrates domain adaptation with GANs and Diffusion models to enhance the quality of low-resolution aerial images.
Our proposed pipeline enhances low-resolution imagery while preserving semantic content, enabling effective tree segmentation without requiring large volumes of manually annotated data. Leveraging models such as pix2pix, Real-ESRGAN, Latent Diffusion, and Stable Diffusion, we generate realistic and structurally consistent synthetic samples that expand the training dataset and unify scale across domains.
This approach not only improves the robustness of segmentation models across different acquisition conditions but also provides a scalable and replicable solution for remote sensing scenarios with scarce annotation resources.
Experimental results demonstrated an improvement of over 50\% in IoU for low-resolution images, highlighting the effectiveness of our method compared to traditional pipelines.
\end{abstract}


\section{Introduction}

Urban forests are increasingly recognized for their significant benefits to human well-being. They contribute to energy savings, reduce stormwater runoff, and improve water quality \citep{velasquez2023implementing, ventura2024individual}. Additionally, these forests provide essential ecosystem services that combat climate change, such as carbon sequestration, oxygen generation, water cycling, soil conservation, and mitigation of the urban heat island effect. Automated tree mapping is essential for effective management of both native and invasive vegetation \citep{lv2023deep, beloiu2023individual}.

For monitoring urban forest resources, satellite remote sensing has been crucial. However, the heterogeneous structure and surface complexity of urban environments, combined with the limited spatial resolution of satellite imagery, pose significant challenges for the accurate detection and delineation of individual trees \citep{velasquez2023implementing, lv2023deep}. In recent years, high-resolution aerial RGB imagery, which is easy to use and available at low cost, has become widely accessible. Unlike satellite images, UAV-acquired imagery typically includes only three RGB channels, which, while providing limited spectral information, enables clear visualization and extraction of structural characteristics such as shape, size, and texture of ground objects \citep{ferreira2020individual}.

In this context, techniques such as semantic segmentation, which offer pixel-based classification, are increasingly employed across a range of applications. Recent advancements in tree detection, classification, and segmentation predominantly utilize deep learning networks, such as ConvNets \citep{ferreira2020individual, iqbal2021coconut, jintasuttisak2022deep}, applied to aerial RGB and multispectral imagery \citep{beloiu2023individual, velasquez2023implementing, ventura2024individual}. More recently, transformers have also been utilized for tree counting in aerial images \citep{chen2022transformer}.

Accurately detecting individual tree from remote sensing data presents a significant challenge for traditional deep learning-based methods due to the variability encountered in cross-regional scenarios \citep{wang2022domain, kapil2024shadowsense, zheng2020cross}. This variability can arise from various factors, including deformations or shifts caused by biased sampling in the spatial domain, changes in acquisition conditions (such as variations in illumination or acquisition angle), or seasonal changes \citep{tuia2021recent}. 

Despite substantial advancements with deep neural networks, their performance improvement largely depends on the availability of extensive labeled training data, which involves costly and labor-intensive data curation \citep{dos2019unsupervised, amirkolaee2024adatreeformer}. The challenge is further compounded when a deep neural network must handle multiple distinct domains. For instance, in tree detecting, each domain might include different scenes (e.g., urban, countryside, farmland), imagery types (e.g., aerial or satellite), and varying levels of tree density, shadows, or overlap among individual trees.

To overcome these challenges, recent works have focused on applying unsupervised domain adaptation in satellite and aerial images. \citet{zheng2020cross} proposed a domain-adaptive method to detect and count cross-regional oil palm trees using an adversarial learning-based multi-level attention mechanism. \citet{wang2022domain} also employed an adversarial domain-adaptive model with a transferable attention mechanism for tree crown detection using high-resolution remote sensing images. More recently, AdaTreeFormer was introduced by \citet{amirkolaee2024adatreeformer}, demonstrating the ongoing trend of combining adversarial learning with attention mechanisms to perform domain adaptation for tree detection in high-resolution images.

To address these limitations, this study introduces an innovative approach that diverges from prior methodologies. Rather than relying solely on adversarial learning and attention mechanisms applied to high-resolution imagery - as commonly seen in recent work - we propose a domain adaptation strategy using image-to-image translation and super-resolution techniques. Our method leverages models such as pix2pix \citep{isola2017image}, Real-ESRGAN \citep{wang2018esrgan}, and both Latent and Stable Diffusion \citep{rombach2022high} to enhance low-resolution aerial images while preserving their semantic integrity. This enables our effective tree segmentation, using SegFormer \citep{xie2021segformer}, without the need for extensive manually labeled datasets, offering a cost-efficient and scalable alternative.

By unifying image scales across domains and automatically generating realistic and annotated synthetic samples, our approach significantly improves model robustness to variations in acquisition conditions, such as flight altitude, tree density, or sensor quality. It provides a versatile and replicable solution for remote sensing scenarios where annotation resources are scarce. Experimental results demonstrate the effectiveness of our pipeline, with IoU improvements exceeding 50\% for low-resolution imagery, clearly outperforming traditional supervised training pipelines.

\section{Methodology}
\label{sec:tree_detection_methodology}

\subsection{Dataset}
\label{sec:tree_detection_dataset}

The images used in the experiments are separated into the datasets $P20$ and $P50$ based on the ground sample distance (GSD) utilized in the capture of the images. The $P20$ dataset consists of $363$ images sized $256 \times 256$ pixels with a 20-centimeter GSD, i.e., each pixel corresponds to approximately 20 cm in the real world. The $P50$ dataset consists of $224$ images sized $256 \times 256$ pixels with a 50-centimeter GSD. Thus, the resolution of the images in the $P20$ dataset is 2.5 times greater than that of the images in the $P50$ dataset.

\begin{table}[htb]
\small
\begin{center}
\def\arraystretch{1.75}
\begin{tabular}{ lccccc }
    \hline
    \hline
      \textbf{Dataset} & \textbf{GSD} & \textbf{Train} & \textbf{Validation} & \textbf{Test} & \textbf{Total}  \\
    \hline
    P20 & 20cm & 218 & 36 & 109 & 363 \\
    P50 & 50cm & 134 & 23 & 67 & 224 \\
    \hline
    \hline
\end{tabular}
\end{center}
\caption[Total of images of train, validation and test sets for datasets $P20$ and $P50$.]{Total of images of train (60\%), validation (10\%) and test (30\%) sets for datasets $P20$ and $P50$ and their respective GSD.}
\label{table:dataset_patches}
\end{table}

\begin{figure}[htb]
\begin{center}
\includegraphics[height=15cm,width=15cm]{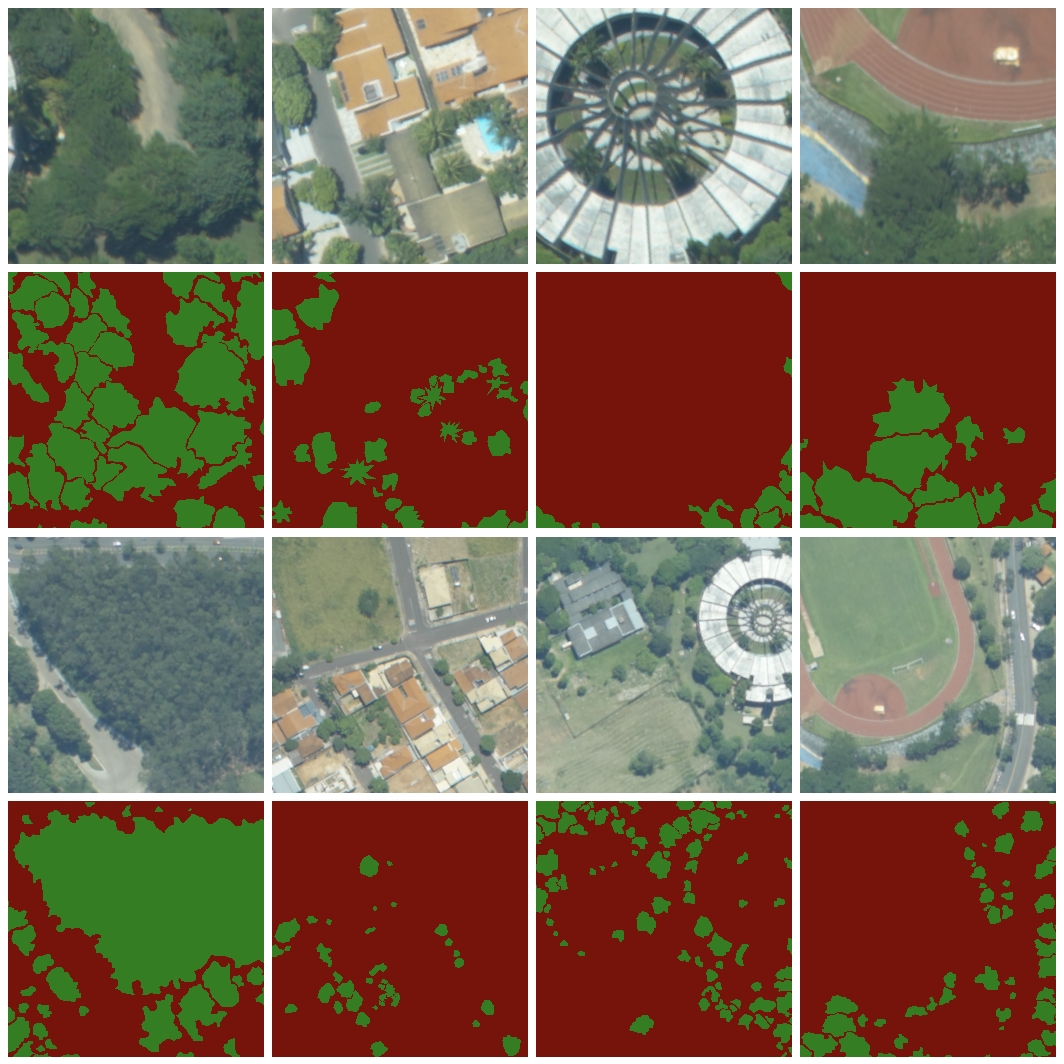}
\end{center}
\caption[Sample images and pixel annotations from datasets $P20$ and $P50$.]{ At the top are sample images from dataset $P20$ with their respective pixel annotations. At the bottom are sample images from dataset $P50$ with their respective pixel annotations. }
\label{figura::patches_dataset}
\end{figure}

The images consist of aerial views of urban environments and have been manually annotated by specialists as either background or tree classes. Sample images from both datasets, along with their respective annotations, can be seen in Figure \ref{figura::patches_dataset}, and the distribution of images in these datasets is shown in Table \ref{table:dataset_patches}.

\begin{figure}[htb]
\begin{center}
\includegraphics[height=5cm,width=16cm]{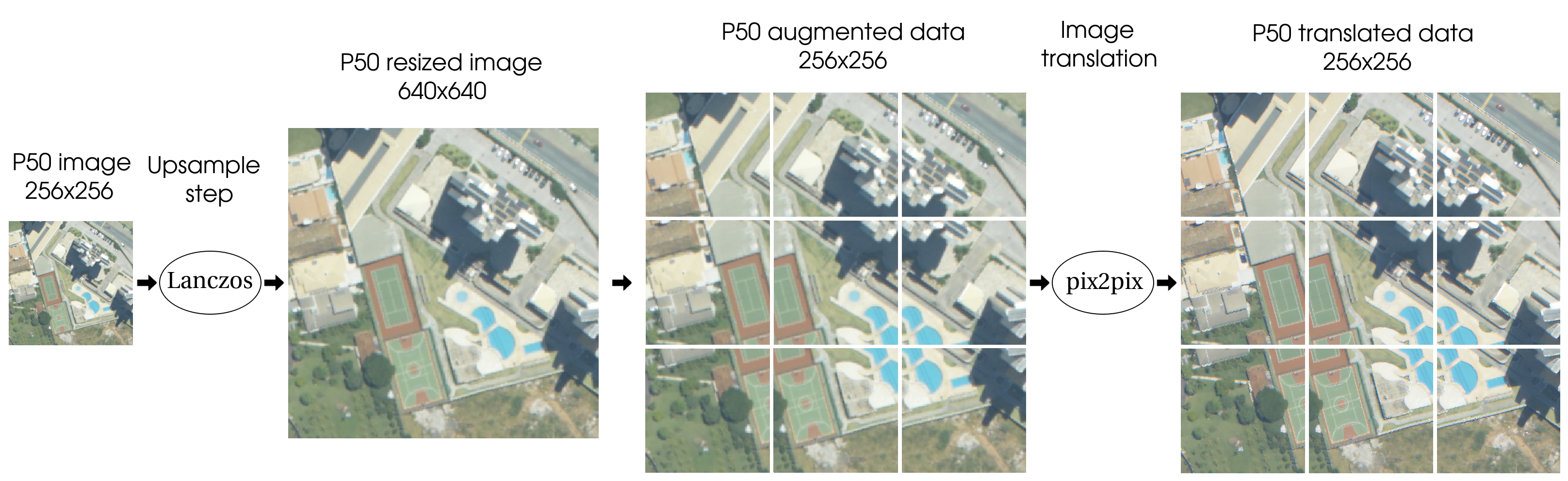}
\end{center}
\caption[Pipeline process using image-to-image translation method.]{ The images of the $P50$ dataset are resized to $640 \times 640$ using Lanczos resampling. For each resized image, we generated 9 patches of size $256 \times 256$ and translated them using pix2pix-trained models. }
\label{figura::pix2pix_pipeline}
\end{figure}

\subsection{Proposed Approach}
\label{sec:tree_detection_upsampling}

\begin{figure}[htb]
\begin{center}
\includegraphics[height=15cm,width=15cm]{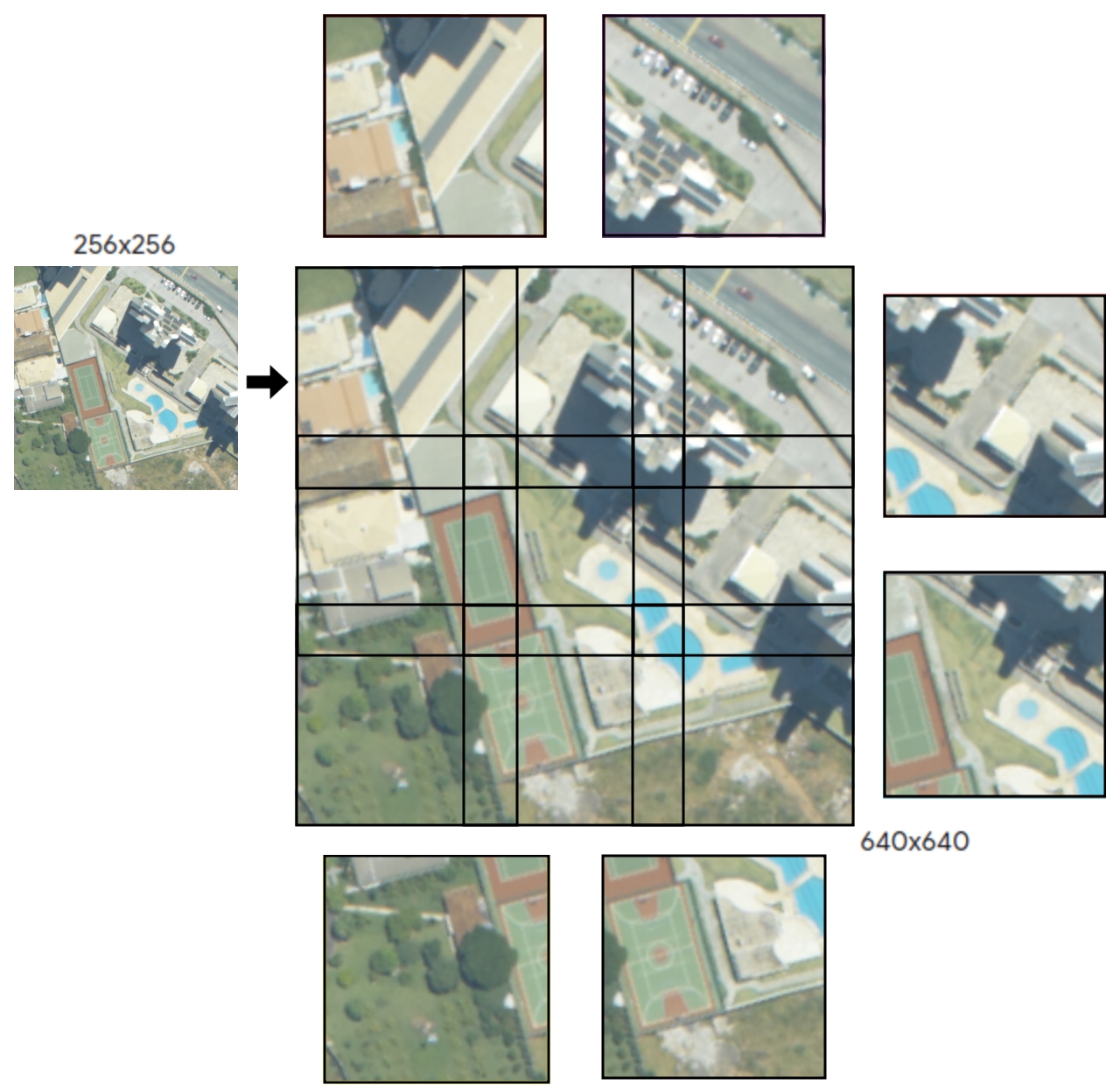}
\end{center}
\caption[Upsample process of images from dataset $P50$.]{ The images of $P50$ dataset are resized from $256 \times 256$ to $ 640 \times 640$ using Lanczos resampling method. After this step, we augmented the data, generating 9 patches of size $256 \times 256$. }
\label{figura::upsample_process}
\end{figure}

Differences in ground sample distance (GSD) across datasets affect the pixel representation of image elements such as trees and roads, as shown in Figure \ref{figura::patches_dataset}. While the size of these elements may remain consistent within a single dataset, variations in GSD between datasets introduce inconsistencies that can hinder model generalization and transferability. To address this, we propose an approach that harmonizes the scale of visual features by adjusting the GSDs through upsampling techniques, ensuring a more uniform representation of key elements across datasets.

We developed two different methods to implement this strategy. In our first method, we upsample the $P50$ dataset, which has a $2.5\times$ difference in centimeters per pixel compared to the $P20$ dataset, by resizing the images from $256 \times 256$ to $640 \times 640$ using the default \textit{ImageMagick} filter, Lanczos resampling \citep{duchon1979lanczos, still2006definitive}, to make the size of objects similar to those in the $P20$ dataset. After this step, we generated 9 patches of size $256 \times 256$.

This process also augments the data in the $P50$ dataset by a factor of 9, increasing it from $224$ images to $2,016$ images. However, this procedure significantly decreases the resolution of these images, which could hamper the performance of network training and increase the data shift compared to the other dataset. To overcome this drawback, we trained pix2pix models to perform image-to-image translation and address the loss of resolution. The pipeline of this method can be seen in Figure \ref{figura::pix2pix_pipeline}. A more detailed visualization of the process for generating patches is illustrated in Figure \ref{figura::upsample_process}.

\begin{figure}[htb]
\begin{center}
\includegraphics[height=6cm,width=13cm]{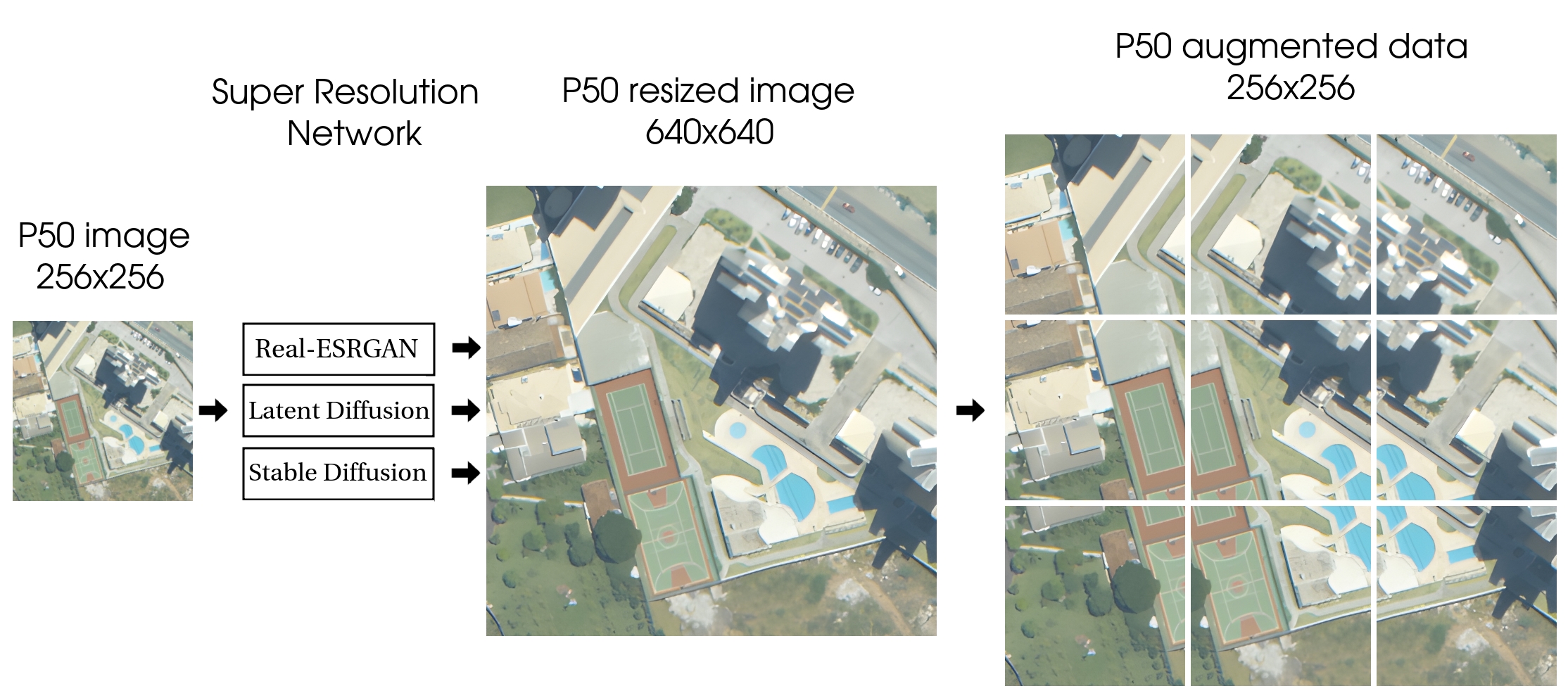}
\end{center}
\caption[Pipeline process using super-resolution models.]{ The images of the $P50$ dataset are resized to $640 \times 640$ using Real-ESRGAN, Latent and Stable Diffusion. For each resized image, we augmented the data, generating 9 patches of size $256 \times 256$.}
\label{figura::sres_pipeline}
\end{figure}

In our second approach, we used recent super-resolution GANs and Diffusion models to upsample the images directly without loss of quality. The pipeline for this method is illustrated in Figure \ref{figura::sres_pipeline}. The advantage of this approach is that we can leverage publicly available models trained on millions of images, unlike the pix2pix model, which needed to be trained from scratch with image pairs generated from our training sets. However, these models do not achieve direct image-to-image translation between the two domains; they primarily enhance resolution to compensate for quality loss during upsampling.

Additionally, it is important to highlight that both approaches used here produce nine times more data from the original images, with these new images having a 2.5 times superior ground sample distance. Since we also updated all annotations for the new GSD automatically, this process helps address the cost of pixel-annotated data and mitigates the drawbacks of low-resolution aerial images. It produces significantly more high-quality annotated data, which is required to train deep learning models efficiently, in a fully automatic way. In the following sections, we provide more details about the methods used.

\subsubsection{Paired Image-to-Image Translation: pix2pix}
\label{sec:tree_detection_dataset_paired}

\begin{figure}[htb]
\begin{center}
\includegraphics[height=15cm,width=15cm]{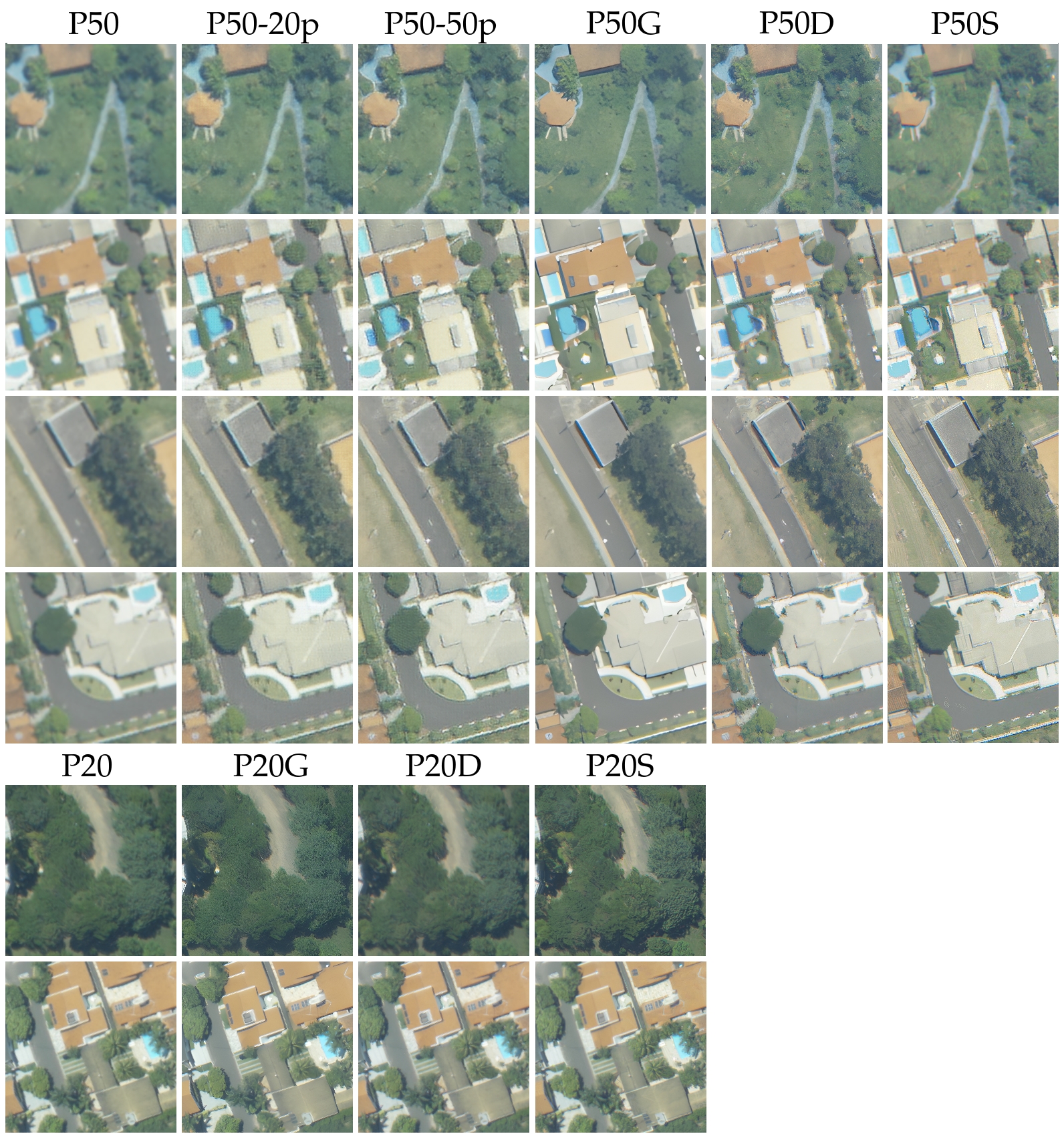}
\end{center}
\caption[Sample images generated using pix2pix, Real-ESRGAN and Diffusion.]{ Sample images generated from datasets $P20$ and $P50$ using pix2pix ($P50-20p$ and $P50-50p$), Real-ESRGAN ($P20G$ and $P50G$), Latent Diffusion ($P20D$ and $P50D$), and Stable Diffusion ($P20S$ and $P50S$). }
\label{figura::generated_images}
\end{figure}

Pix2pix is an image-to-image translation GAN and has shown promising results in datasets with a paired image relationship between the source and target domains, such as the Facade and Cityscapes  datasets \citep{tylevcek2013spatial, cordts2016cityscapes}. The image-to-image translation used here could alleviate distortions in the generated images that might otherwise decrease the segmentation performance in subsequent steps. However, since we lack a direct relationship between the images of the two datasets, $P20$ and $P50$, to perform a true paired translation, we proposed two approximate mapping approaches.

\begin{table}[htb]
\small
\begin{center}
\def\arraystretch{1.75}
\begin{tabular}{ lcccccc }
    \hline
    \hline
      \textbf{Dataset} & \textbf{Generation Method} & \textbf{GSD} & \textbf{Train} & \textbf{Validation} & \textbf{Test} & \textbf{Total} \\
    \hline
    P50-20p  & \makecell{pix2pix trained \\ with $P20$ pairs} & 20cm & 1206 & 207 & 603 & 2016 \\
    P50-50p  & \makecell{pix2pix trained \\ with $P50$ pairs} & 20cm & 1206 & 207 & 603 & 2016 \\
    \hline
    \hline
\end{tabular}
\end{center}
\caption[Total of images of train, validation and test sets for the datasets generated using pix2pix translation.]{Total of images of train (60\%), validation (10\%) and test (30\%) sets for the datasets generated using pix2pix translation. Image pairs used in the training of $P50$-20p can be seen in Figure \ref{figura::pix2pix_pairs_20}, and those used in the training of $P50$-50p can be seen in Figure \ref{figura::pix2pix_pairs_50}. }
\label{table:dataset_p50pix}
\end{table}

To perform the mapping required for paired image-to-image translation used in pix2pix, we reduced the resolution of the images in datasets $P20$ and $P50$. For dataset $P20$, we used resolutions of $32 \times 32$, $64 \times 64$, $96 \times 96$, $128 \times 128$, and $192 \times 192$.  For dataset $P50$, we used resolutions of $16 \times 16$, $32 \times 32$, $64 \times 64$, $96 \times 96$, and $128 \times 128$.  After resizing to these smaller resolutions, we upscaled the images back to $256 \times 256$ without any preprocessing steps and generated paired images. Examples of these paired images can be found in \textit{Supplementary Material}. We trained two different pix2pix models using these pairs.

We used the $2,016$ images obtained after applying the Lanczos method to the $P50$ dataset, as illustrated in Figure \ref{figura::upsample_process}, as input for the pix2pix models, generating two new datasets: $P50-20p$ and $P50-50p$. The distribution of images in these datasets is described in Table \ref{table:dataset_p50pix}. Sample images from these datasets are shown in Figure \ref{figura::generated_images}.

\subsubsection{Super-Resolution Models: Real-ESRGAN, Latent and Stable Diffusion}
\label{sec:tree_detection_dataset_sres}

\begin{figure}[htb]
\begin{center}
\includegraphics[height=15cm,width=15cm]{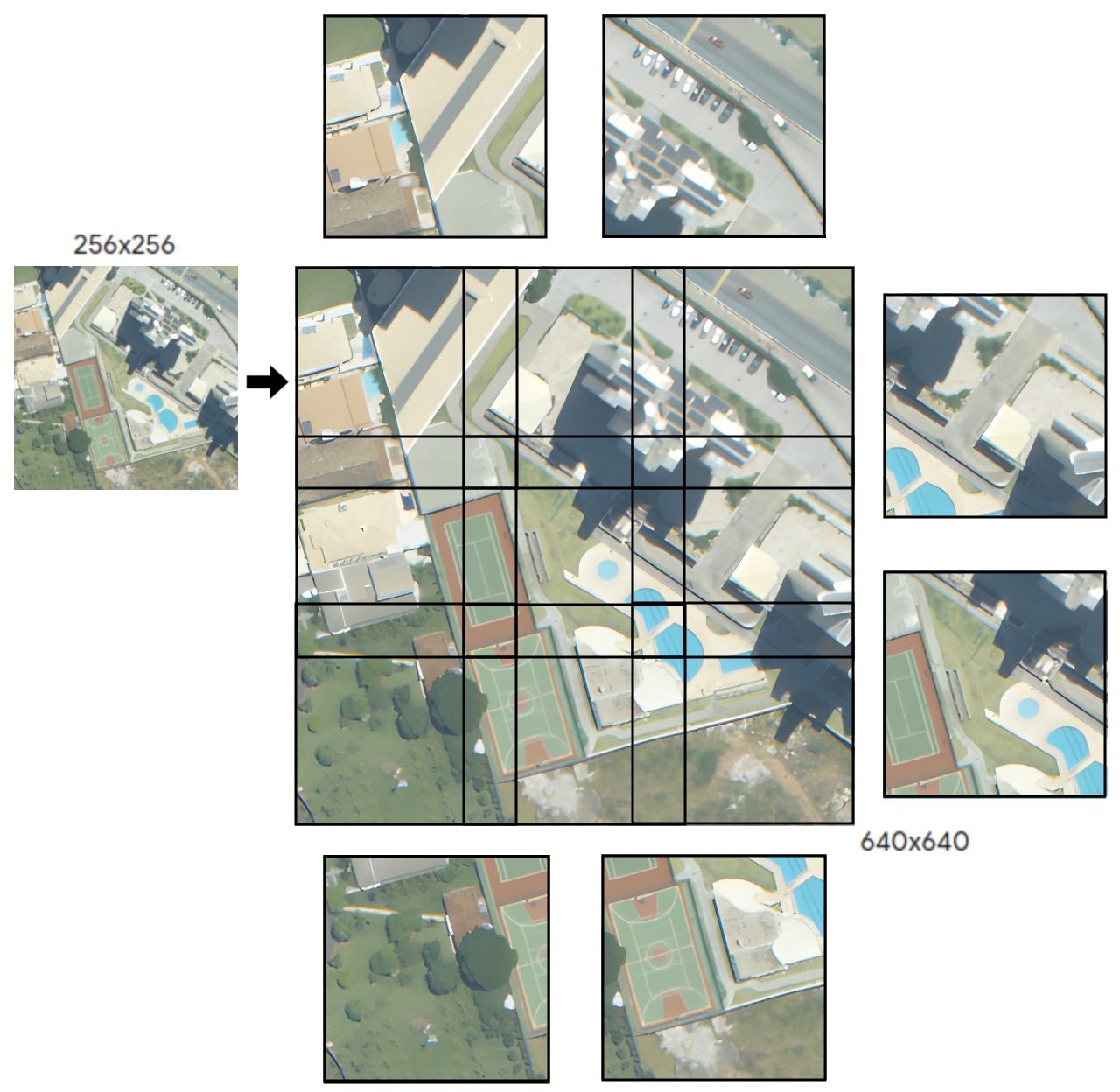}
\end{center}
\caption[Upsample process of images from dataset $P50$ using Real-ESRGAN.]{ The images of $P50$ dataset are upscaled from $256 \times 256$ to $ 640 \times 640$ using Real-ESRGAN. After this step, we generated 9 patches of size $256 \times 256$. The visual quality is significantly better compared to the resized images shown in Figure \ref{figura::upsample_process}.}
\label{figura::upsample_process_gan}
\end{figure}

\begin{table}[htb]
\small
\begin{center}
\def\arraystretch{1.75}
\begin{tabular}{ lcccccc }
    \hline
    \hline
      \textbf{Dataset} & \textbf{Generation Method} & \textbf{GSD} & \textbf{Train} & \textbf{Validation} & \textbf{Test} & \textbf{Total} \\
    \hline
    P20G & Real-ESRGAN & 20cm & 218 & 36 & 109 & 363 \\
    P50G & Real-ESRGAN & 20cm &1206 & 207 & 603 & 2016 \\
    P20D & Latent Diffusion & 20cm &218 & 36 & 109 & 363 \\
    P50D & Latent Diffusion & 20cm &1206 & 207 & 603 & 2016 \\
    P20S & Stable Diffusion & 20cm &218 & 36 & 109 & 363 \\
    P50S & Stable Diffusion & 20cm &1206 & 207 & 603 & 2016 \\
    \hline
    \hline
\end{tabular}
\end{center}
\caption[Total of images of train, validation and test sets for each super-resolution dataset.]{Total of images of train (60\%), validation (10\%) and test (30\%) sets for each super-resolution dataset. }
\label{table:dataset_sres}
\end{table}

We used the Real-ESRGAN and Diffusion public models, without any fine-tuning, to generate our $640 \times 640$ images from dataset $P50$. Using the resulting super-resolution images, we generated 9 patches of size $256 \times 256$, as described in Figure \ref{figura::upsample_process_gan}. For dataset $P20$, we upscaled the original images to $640 \times 640$ using the models and then resized them back to the original size of $256 \times 256$ to maintain similarity with the images generated by the previous pipeline. The distribution of images in each generated dataset can be seen in Table \ref{table:dataset_sres}, where the suffix G represents Real-ESRGAN, the suffix D represents Latent Diffusion, and the suffix S represents Stable Diffusion.

Using Stable Diffusion, we have the option to provide a prompt that guides the image generation. While this could be an advantage over Latent Diffusion, for this work, this feature poses a challenge in choosing a prompt that optimizes our segmentation results. Since evaluating the optimal prompt for the segmentation task is somewhat beyond the scope of this work, we used the prompt \textit{Enhance the resolution of this aerial city image without applying any filter}, which was selected from a few alternatives based on its superior qualitative visual results.

\subsection{Evaluation Metrics}
\label{sec:tree_detection_metrics}

To assess and compare the networks evaluated in the experiments, we used the metric commonly applied in the literature: intersection over union (IoU) at the pixel level, described in Equation \ref{formula:iou_tree}.

\begin{equation} \label{formula:iou_tree}
\begin{split}
    IoU &= \frac{P \cap GT}{P \cup GT},
\end{split}
\end{equation}
where P corresponds to model prediction and GT corresponds to Ground Truth. 

In all experimental results presented here, the notation $P_S \rightarrow P_T$ indicates that the model was trained on images from dataset $P_S$ and evaluated on test images from dataset $P_T$. Thus, $S = T$ signifies that training and test images come from the same dataset, while $S \neq T$ denotes a scenario where the model is trained on one dataset and evaluated on a different dataset.

\subsection{Experimental Setup}
\label{sec:tree_detection_setup}

We ran our experiments with SegFormer, pix2pix, and Real-ESRGAN using the free version of Google Colab with a T4 GPU. For experiments with Latent Diffusion, and Stable Diffusion, we utilized an Intel(R) Core (TM) i7-5820K CPU @ 3.30GHz with 32 GB of RAM, and an Nvidia GeForce GTX TITAN X GPU with 12 GB GDDR5 memory and 3072 CUDA Cores.

In our supervised segmentation tests with SegFormer, we utilized the available architectures in MMSegmentation, accessible at \href{https://github.com/open-mmlab/mmsegmentation}{https://github.com/open-mmlab/mmsegmentation}. For training, we used the base configuration files provided by MMSegmentation, specifically using the Cityscapes configuration with the MIT-B5 backbone, a crop size of $1024 \times 1024$, and a learning rate schedule set at $160000$. Additionally, we adjusted the image scale to $256 \times 256$, modified the number of classes in the decode/auxiliary head to $2$, and resized the crop size to $128 \times 128$ to better suit our dataset.

For pix2pix training, we utilized the original code provided by the authors, accessible at \href{https://github.com/junyanz/pytorch-CycleGAN-and-pix2pix}{github.com/junyanz/pytorch-CycleGAN-and-pix2pix}. Each model was trained for 200 epochs with decay initiated after 100 epochs. No additional training or fine-tuning was conducted for Real-ESRGAN. Inference was performed using the default configurations provided in the script available from the authors' repository at \href{https://github.com/xinntao/Real-ESRGAN.git}{github.com/xinntao/Real-ESRGAN.git}.

For Latent and Stable Diffusion, we utilized the implementation provided by the authors in python library format, accessible at \href{https://github.com/CompVis/latent-diffusion}{github.com/CompVis/latent-diffusion} and \href{https://github.com/CompVis/stable-diffusion}{github.com/CompVis/stable-diffusion}. The images resulting from inference by the GANs and Diffusion models were used to train the SegFormer model.

Unlike Real-ESRGAN, the outscale parameter of the pre-trained Diffusion models could not be adjusted to a value smaller than 4. Due to our machine's 12GB memory limitation, we were unable to resize images from $256 \times 256$ to $1024 \times 1024$ directly. Therefore, we divided our original images into 4 patches of $128 \times 128$, upscaled them using the Diffusion models, and then used the 4 upscaled patches to reconstruct the image with size $1024 \times 1024$. We acknowledge that this step could have impacted our results and consider this aspect a limitation of the Diffusion pre-trained models.

\section{Results and Discussion}

\subsection{Baseline}
\label{sec:tree_detection_src_only}

We evaluated the performance of supervised segmentation using SegFormer on two original datasets, $P20$ and $P50$, without upsampling the original images. The results are presented on the left side of Table \ref{table:patches_supervised}. While both datasets achieved considerable performance in terms of IoU metric, dataset $P20$ exhibited a higher IoU than dataset $P50$. This outcome was anticipated, given that dataset $P20$ comprises higher-resolution images and a larger training set.

\begin{table}[htb]
\small
\begin{center}
\def\arraystretch{1.75}
\begin{tabular}{ lcc|cc }
    \hline
    \hline
      & \textbf{P20 $\rightarrow$ P20} & \textbf{P50 $\rightarrow$ P50} & \textbf{P50 $\rightarrow$ P20} & \textbf{P20 $\rightarrow$ P50} \\
    \hline
    \multicolumn{5}{l}{\textbf{SegFormer  (MiT-B5)}} \\
    \hline
    \text{Background} & 94.87 & 95.56 & 91.05 & 94.22 \\
    \text{Trees} & \textbf{77.44} & 70.18 & 57.43 & 63.27 \\
    \text{Average} & 86.15 & 82.87 & 74.25 & 78.75 \\
    \hline
    \hline
\end{tabular}
\end{center}
\caption[IoU of supervised training using the original datasets.]{ IoU of supervised training using the original datasets. On the right side, the source model only results are shown. In bold, the best result for the Trees class.}
\label{table:patches_supervised}
\end{table}

We also evaluated the models on a different dataset than those used for training (i.e., source model only). The results are presented on the right side of Table \ref{table:patches_supervised}. When segmenting target images with models trained on images from a different domain, a noticeable decrease in IoU is observed due to data shift. This performance drop is particularly pronounced when using the model trained on dataset $P50$ to segment images from dataset $P20$, where the IoU decreases from 77.44 to 57.43 for the Trees class, approximately a 25.8\% drop.

\begin{figure}
\begin{center}
\includegraphics[height=21.5cm,width=10.5cm]{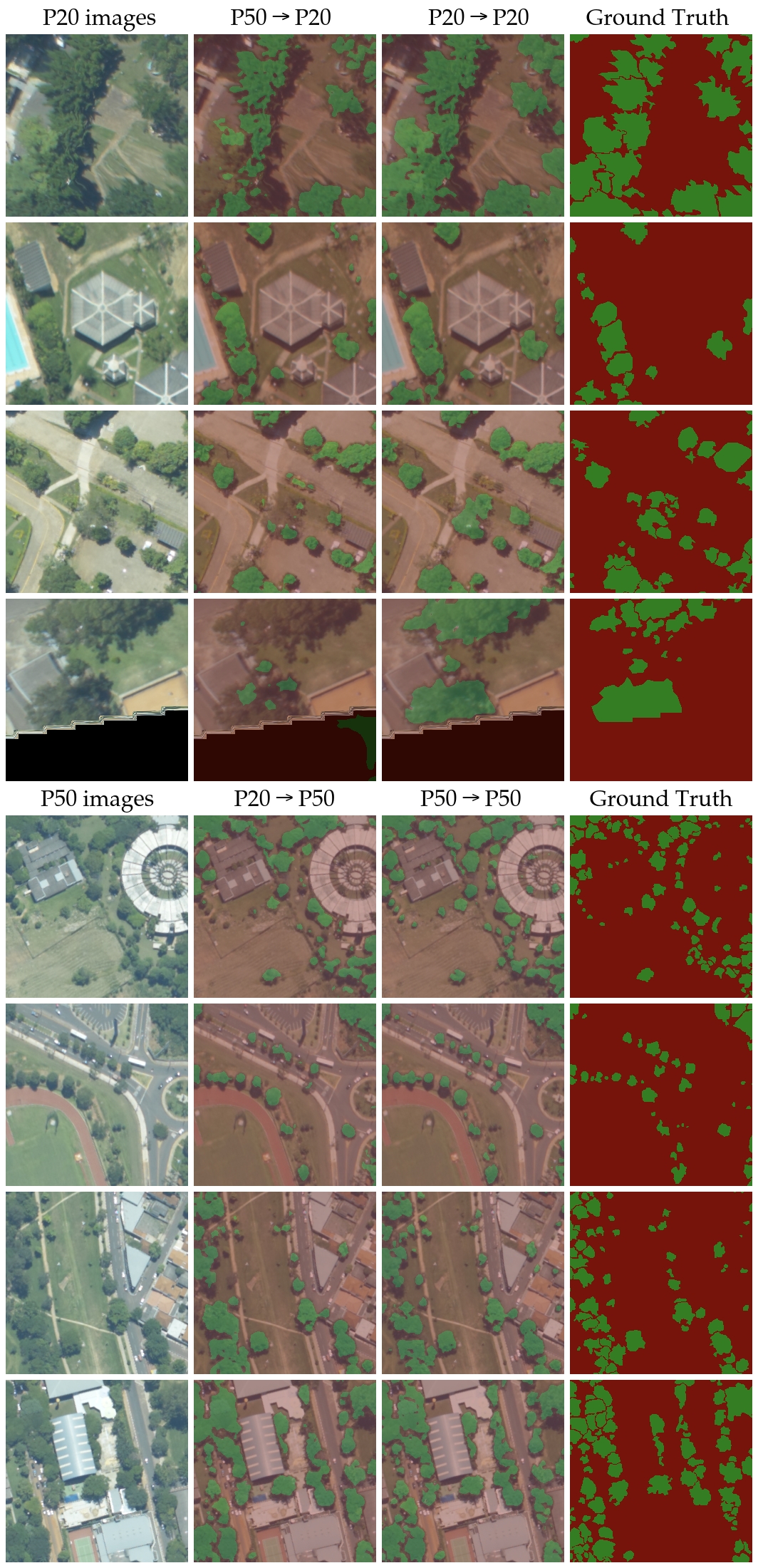}
\end{center}
\caption[Predictions using the SegFormer model trained with images from datasets $P20$ and $P50$.]{ Predictions using the SegFormer model trained with images from datasets $P20$ and $P50$.  The models performed well even when segmenting images from a different domain.}
\label{figura::pred_original}
\end{figure}

In Figure \ref{figura::pred_original} we can see the visual predictions using the SegFormer model trained with images from datasets $P20$ and $P50$. The models performed well even when segmenting images from a different domain. However, the $P50$ model failed to detect some large trees and occasionally misidentified grass as trees in the $P20$ images. The $P20$ model failed to detect smaller trees in the $P50$ images, but the reduced size of the trees generated a smaller impact on the average IoU.

Although we can consider the performance of the source model only reasonable in these experiments, given the similarity of the images in both datasets, the next sections analyze techniques aimed at improving these results, as well as enhancing the performance of supervised segmentation.

\subsection{Paired Image-to-Image Translation}
\label{sec:tree_detection_paired}

\subsubsection{pix2pix}

\begin{table}[htb]
\small
\begin{center}
\def\arraystretch{1.75}
\begin{tabular}{ lccc }
    \hline
    \hline
      & \textbf{P50-20p$\rightarrow$P50-20p } & \textbf{P50-50p$\rightarrow$P50-50p } & \textbf{P50$\rightarrow$P50} \\
    \hline
    \multicolumn{4}{l}{\textbf{SegFormer  (MiT-B5)}} \\
    \hline
    \text{Background} & 96.05 & 95.99 & 95.56 \\
    \text{Trees} & \textbf{73.25} & 72.77 & 70.18  \\
    \text{Average} & 84.65 & 84.37 & 82.87 \\
    \hline
    \hline
\end{tabular}
\end{center}
\caption[IoU of supervised training with images generated by the pix2pix models.]{ IoU of supervised training with images generated by the pix2pix models. compared to the original datasets. In bold, the best result for the Trees class.}
\label{table:patches_supervised_pix2pix}
\end{table}

We trained two pix2pix models using the pairs described in Section \ref{sec:tree_detection_dataset_paired}. These models were used to generate two new datasets, $P50-20p$ and $P50-50p$, which consist of translated images from dataset $P50$ after applying the upsampling process. The results of the SegFormer supervised segmentation trained with these models can be seen in Table \ref{table:patches_supervised_pix2pix}. In both cases, we observe an improvement in IoU compared to supervised segmentation using the original images. 

\begin{figure}[htb]
\begin{center}
\includegraphics[height=13.5cm,width=16cm]{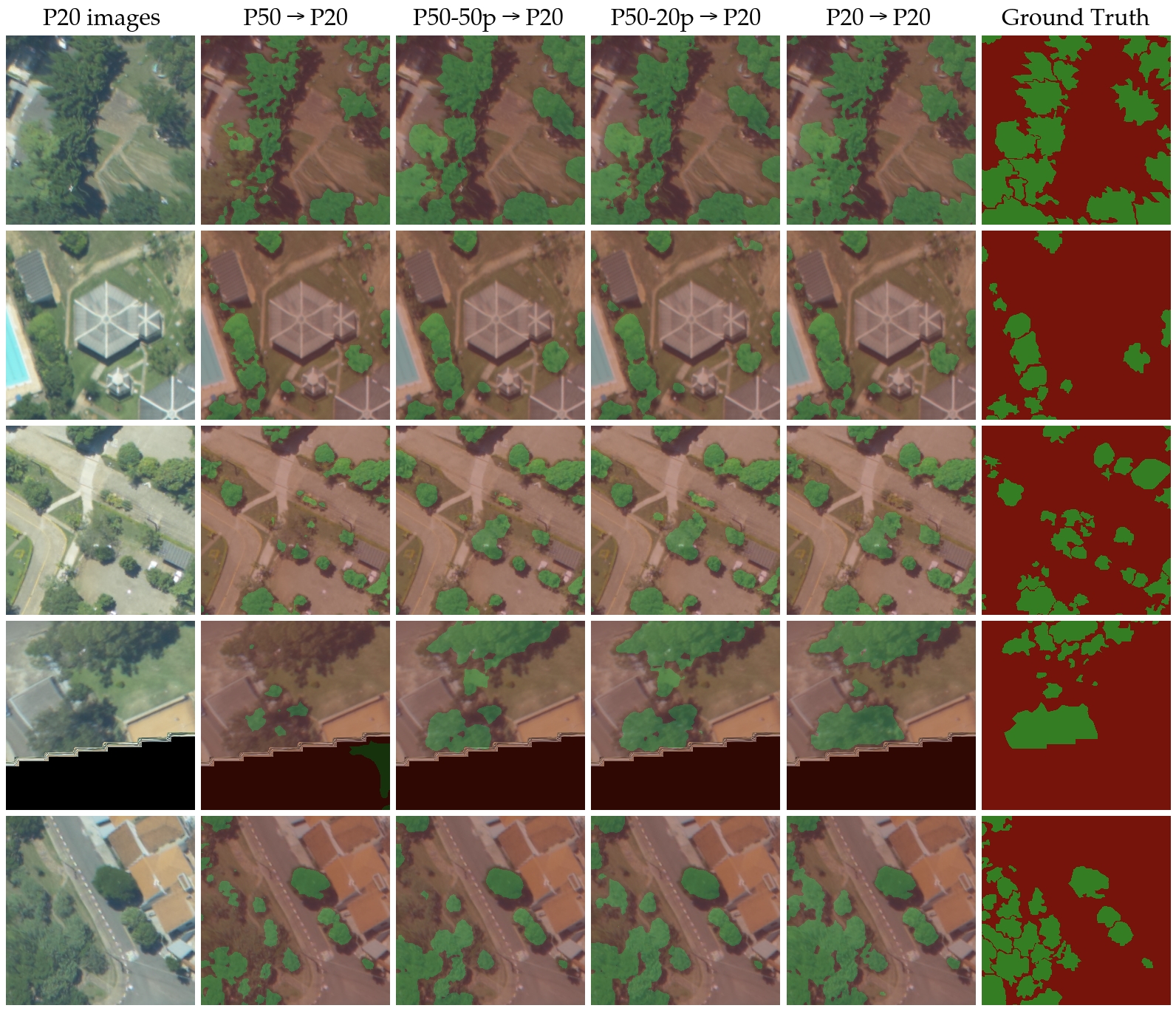}
\end{center}
\caption[Predictions using the SegFormer model trained with images from datasets $P50-20p$ and $P50-50p$.]{ Predictions using the SegFormer model trained with images from datasets $P50-20p$ and $P50-50p$ in $P20$ images.  In the bottom left corner of the first and last images, we can see the improvement of the pix2pix models in detecting larger trees.}
\label{figura::pred_pix2pix}
\end{figure}

\begin{table}[htb]
\small
\begin{center}
\setlength{\tabcolsep}{0.1em}
\def\arraystretch{1.75}
\begin{tabular}{ l|ccc|ccc }
    \hline
    \hline
      & \textbf{P20$\rightarrow$P50-20p } & \textbf{P20$\rightarrow$P50-50p } & \textbf{P20$\rightarrow$P50 } & \textbf{P50-20p$\rightarrow$P20 } & \textbf{P50-50p$\rightarrow$P20 } & \textbf{P50$\rightarrow$P20} \\
    \hline
    \multicolumn{7}{l}{\textbf{SegFormer  (MiT-B5)}} \\
    \hline
    \text{Background} & 94.65 & 94.48 & 94.22 & 93.07 & 92.94 & 91.05  \\
    \text{Trees} & \textbf{67.20} & 66.29 & 63.27 & \textbf{68.05} & 67.43 & 57.43  \\
    \text{Average} & 80.92 & 80.38 & 78.75 & 80.56 & 80.19 & 74.25 \\
    \hline
    \hline
\end{tabular}
\end{center}
\caption[IoU of the src-only evaluation with images generated by the pix2pix models.]{ IoU of the src-only evaluation with images generated by the pix2pix models compared to the original datasets. In bold, the best results for the Trees class.}
\label{table:patches_src_only_pix2pix}
\end{table}

However, it is important to highlight that we are not evaluating the translated images from dataset $P50$ directly but rather the corresponding augmented data generated through the upsampling process; thus, this improvement could also be attributed to the data augmentation process.

Nevertheless, it is an interesting finding that, in these experiments, we were able to enhance our segmentation results using the same network, SegFormer, without the need for more labeled images for training. Instead, we achieved this increase by generating more images at the same size but with lower resolution and then improving the quality using paired image-to-image translation, showing the potential of our data augmentation method.

We also evaluated these models as source model only on the test images from dataset $P20$ and evaluated the model trained with images from dataset $P20$ on the images generated by pix2pix. The results can be seen at Table \ref{table:patches_src_only_pix2pix}. In all tests, we achieved significant improvements compared to the results on the original images of dataset $P50$ without using image-to-image translation. The best model trained with pix2pix images improved the IoU for the Trees class from 57.43 to 68.05, reducing the gap with the supervised results of  $P20 \rightarrow P20$, 77.43, by approximately 60\%.

\subsection{Super-Resolution Models}
\label{sec:tree_detection_srs}

We used the super-resolution models to generate high-resolution images from the datasets $P20$ and $P50$, as described in Section \ref{sec:tree_detection_dataset_sres}. We evaluated the SegFormer model trained on these images and compared its performance to training using the original images. The results for each network evaluated are detailed in the following sections.

\subsubsection{Real-ESRGAN}

\begin{table}[htb]
\small
\begin{center}
\def\arraystretch{1.75}
\begin{tabular}{ l|cc|cc }
    \hline
    \hline
      & \textbf{P20G $\rightarrow$ P50G} & \textbf{P20 $\rightarrow$ P50} & \textbf{P50G $\rightarrow$ P20G} & \textbf{P50 $\rightarrow$ P20} \\
    \hline
    \multicolumn{5}{l}{\textbf{SegFormer  (MiT-B5)}} \\
    \hline
    \text{Background} & 94.86 & 94.22 & 92.45 & 91.05  \\
    \text{Trees} & \textbf{66.57} & 63.27 & \textbf{63.92} & 57.43  \\
    \text{Average} & 80.71 & 78.75 & 78.19 & 74.25 \\
    \hline
    \hline
\end{tabular}
\end{center}
\caption[IoU of the src-only evaluation with images upscaled using Real-ESRGAN.]{ IoU of the src-only evaluation with images upscaled using Real-ESRGAN, compared to the original datasets. In bold, the best results for the Trees class.}
\label{table:patches_src_only_gan}
\end{table}

Although the images generated by Real-ESRGAN exhibit superior visual quality compared to those generated by pix2pix models, as depicted in Figure \ref{figura::generated_images}, the results of our experiments were slightly inferior to those achieved by SegFormer trained with images translated by pix2pix models, as shown in Table \ref{table:patches_src_only_gan}. This difference can be attributed to the fact that while we trained the pix2pix models using images from our specific datasets, Real-ESRGAN uses a super-resolution model trained on general images.

This lack of training could have led the network to distort the semantic information of some pixels, resulting in a decrease in the segmentation results. However, it is worth highlighting that omitting the training step sped up our pipeline. Moreover, while semantic distortion of pixels can significantly impact segmentation tasks, in other tasks such as object detection, this effect is generally negligible.

\subsubsection{Latent and Stable Diffusion}

\begin{table}[htb]
\small
\begin{center}
\setlength{\tabcolsep}{0.1em}
\def\arraystretch{1.75}
\begin{tabular}{ l|ccc|ccc }
    \hline
    \hline
      & \textbf{P20D$\rightarrow$P50D } & \textbf{P20S$\rightarrow$P50S } & \textbf{P20$\rightarrow$P50 } & \textbf{P50D$\rightarrow$P20D } & \textbf{P50S$\rightarrow$P20S } & \textbf{P50$\rightarrow$P20} \\
    \hline
    \multicolumn{7}{l}{\textbf{SegFormer  (MiT-B5)}} \\
    \hline
    \text{Background} & 94.42 & 94.63  & 94.22 & 92.59 & 91.63 & 91.05  \\
    \text{Trees} & 65.58 & \textbf{65.59} & 63.27 & \textbf{65.36} & 62.73 & 57.43  \\
    \text{Average} & 80.00 & 80.11 & 78.75 & 78.97 & 77.18 & 74.25 \\
    \hline
    \hline
\end{tabular}
\end{center}
\caption[IoU of the src-only evaluation with images upscaled using Latent and Stable Diffusion.]{ IoU of the src-only evaluation with images upscaled using Latent and Stable Diffusion, compared to the original datasets. In bold, the best results for the Trees class.}
\label{table:patches_src_only_diffusion}
\end{table}

\begin{figure}[htb]
\begin{center}
\includegraphics[height=13.5cm,width=16cm]{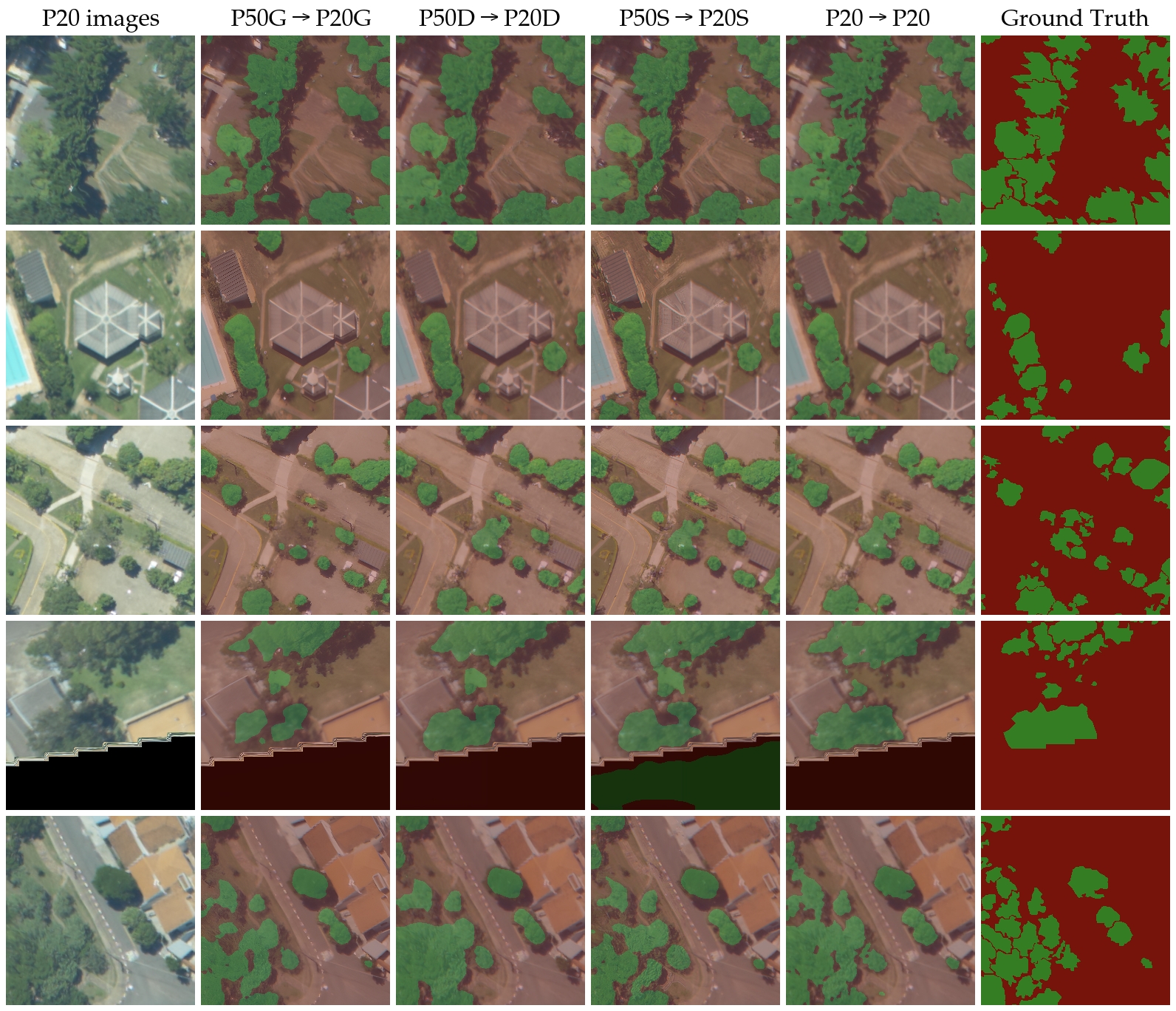}
\end{center}
\caption[Predictions using the SegFormer model trained with images from datasets $P50G$, $P50D$, and $P50S$,.]{ \small{ Latent diffusion produces better segmentation results than ESRGAN, despite the GAN model generating images with better visual quality. Ironically, Stable Diffusion suffers from instability in the fourth image, a behavior that may have been influenced by prompt usage.} }
\label{figura::pred_sres}
\end{figure}

With our Diffusion models, we obtained results similar to Real-ESRGAN, as shown in Table \ref{table:patches_src_only_diffusion}. We also experimented a combination of models trained using Latent and Stable Diffusion. One interesting finding was that our best results were achieved using a model trained with images from dataset $P50D$ to segment the test images from dataset $P20S$, achieving an IoU of 67.79 for the Trees class, superior to our results shown in the Table.

However, it's difficult to establish a specific reason for this behavior, mainly due to the fact that the resulting images from Stable Diffusion are strongly influenced by the prompt used. Nevertheless, this aspect may highlight the possibilities that can be explored with the use of Stable Diffusion in similar tasks. In Figure \ref{figura::pred_sres}, we can observe a visual comparison of the segmentation results of datasets generated by the super-resolution methods.

\subsection{Low Resolution Images}
\label{sec:tree_detection_low}

Despite a 2.5-fold resolution difference between our original datasets $P20$ and $P50$, the visual quality in both cases was good, and the slight disparity in resolution between the datasets allowed us to achieve satisfactory results with the source model only approach, even without applying image translation or using super-resolution networks. One scenario not addressed in our experiments with these datasets is using our trained models with images of lower quality than those used in training.

\begin{table}[htb]
\small
\begin{center}
\def\arraystretch{1.75}
\begin{tabular}{ lcccc|c }
    \hline
    \hline
      & \textbf{P20 $\rightarrow$ P20lr} & \textbf{P20 $\rightarrow$ P20lp} & \textbf{P20 $\rightarrow$ P20lG} & \textbf{P20 $\rightarrow$ P20lD} & \textbf{P20 $\rightarrow$ P20} \\
    \hline
    \multicolumn{6}{l}{\textbf{SegFormer  (MiT-B5)}} \\
    \hline
    \text{Background} & 89.72 & 92.43 & 90.22  & 90.41 & 94.87 \\
    \text{Trees} & 50.99 & \textbf{67.80}  & 61.71 &  61.60  & \textbf{77.44} \\
    \text{Average} & 70.36 & 80.11 & 75.97 & 76.00 & 86.15 \\
    \hline
    \hline
\end{tabular}
\end{center}
\caption[IoU of the src-only evaluation of SegFormer model trained with images from datasets $P20$ in upscaled images.]{ IoU of the src-only evaluation using the model trained with images from datasets $P20$ against low resolution and upscaled images using pix2pix, Real-ESRGAN, Latent Diffusion, and Stable Diffusion. In bold, the best result for the Trees class.}
\label{table:patches_src_only_lowres}
\end{table}

We decided to simulate this scenario to evaluate the performance of the techniques presented here in enhancing the quality of low-resolution images. To simulate it, we resized the original $ 256 \times 256$ images from the $P20$ dataset to $32 \times 32$, decreasing their resolution by 8 times. This represents a difference significantly greater than the 2.5 times difference in our datasets.

\begin{figure}[htb]
\begin{center}
\includegraphics[height=13.5cm,width=16cm]{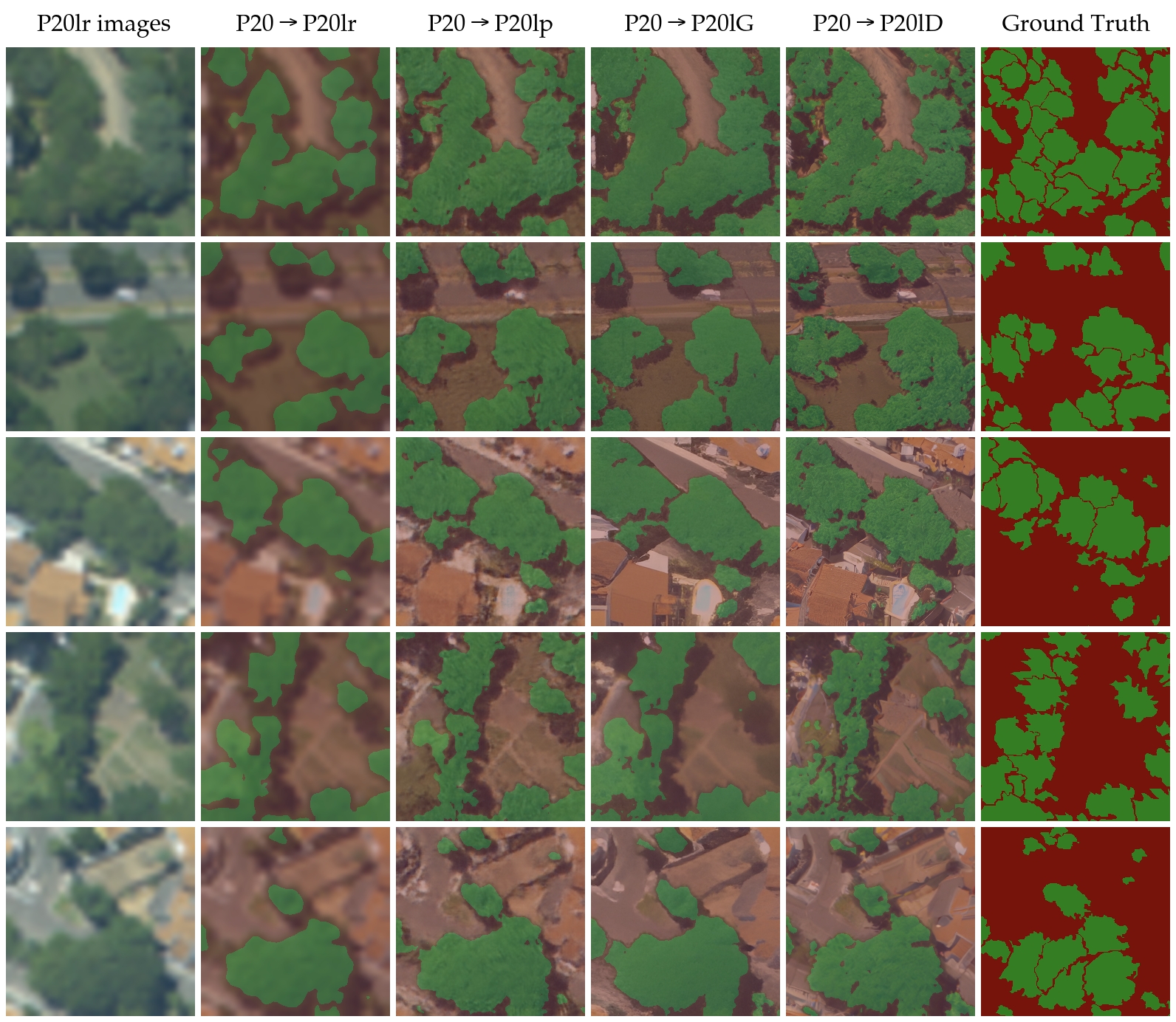}
\end{center}
\caption[Predictions using the SegFormer model trained with images from datasets $P20$ in upscaled images.]{ Predictions using the SegFormer model, trained with original images from dataset $P20$, in the low resolution images and their respective upscaled images using pix2pix, Real-ESRGAN, and Latent Diffusion.}
\label{figura::pred_lowres}
\end{figure}

Through this process, we created the dataset $P20$lr (\textit{P20 low resolution}) and used it to test our GANs and Diffusion methods, creating new datasets with translated images. We generated the dataset $P20$lp after applying pix2pix translation in $P20$lr, the dataset $P20$lG after increasing the resolution using Real-ESRGAN, and the dataset $P20$lD after enhancing the resolution with Latent Diffusion. Examples of images from these datasets can be found in \textit{Supplementary Material}.

In Table \ref{table:patches_src_only_lowres}, we present the IoU results of segmentation using our model trained with images from dataset $P20$. There is a noticeable decrease in performance when our model trained with original $P20$ images segments low-resolution images from database $P20$lr. However, when segmenting target images translated by the pix2pix model, this same model achieved significantly better results compared to those obtained using super-resolution models, despite the visually superior quality of images generated by Latent Diffusion, particularly evident in the depiction of roofs.

This evaluation corroborates the idea that, for the approach used in this work, preserving the semantic information of original pixels is more crucial for segmentation results than achieving high visual quality in the generated images. However, it is important to acknowledge the capability of super-resolution models to generate coherent images from low-resolution inputs using a publicly available checkpoint without fine-tuning and the training process required by pix2pix models. The visual predictions, compared to the ground truth, can be seen in Figure \ref{figura::pred_lowres}.

\section{Conclusion}
\label{sec:tree_detection_conclusion}

In this work, we introduced an approach to enhance the resolution of aerial images to improve tree detection performance by utilizing image-to-image translation and super-resolution methods. Our method introduced a novel data augmentation technique, employing upsampling to generate high-quality annotated samples with varying ground sample distances (GSD). This approach also addresses the costly and labor-intensive process of manually labeling data. 

Our data augmentation pipeline, which combines upsampling with translation and super-resolution steps, can be applied with different scaling factors to create new labeled images across a range of GSDs. This process enables the network to adapt to different image capture heights, thereby increasing the robustness of the supervised model when applied to new domains. Our evaluation revealed that lightweight models, such as pix2pix, can compete effectively with more recent and complex networks in translating images when trained appropriately.

In addition, we also conducted experiments reducing the resolution of our original dataset images, which were generally of high quality, by a factor of eight and evaluated the model's performance on both the original and enhanced images. The results demonstrated that our upsampling pipeline using pix2pix improved IoU tree detection performance by more than 50\% when compared to the low-resolution images, validating the effectiveness of our upsampling strategy. The methods for enhancing resolution presented in this work can be applied in scenarios where remote sensing images lack the necessary quality for achieving high accuracy in computer vision tasks, such as detection, classification, and segmentation.

\bibliographystyle{plainnat}  
\bibliography{references} 

\clearpage

\appendix
\section{Supplementary Material}

\begin{figure}[htb]
\begin{center}
\includegraphics[height=7.5cm,width=15cm]{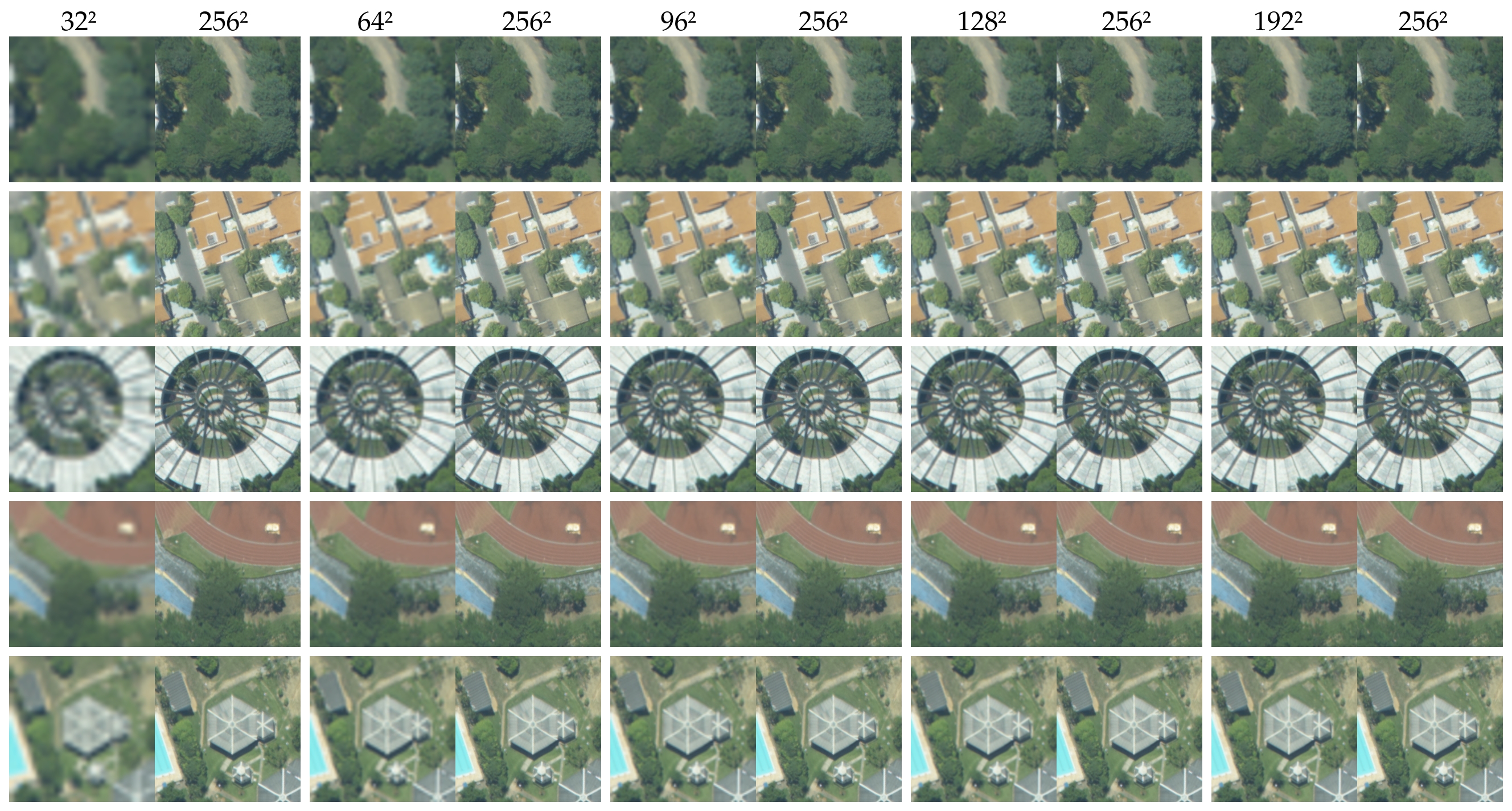}
\end{center}
\caption[Pix2pix training pairs with images of the $P20$ dataset.]{ Pix2pix training pairs with images of the $P20$ dataset at resolutions of $32 \times 32$, $64 \times 64$, $96 \times 96$, $128 \times 128$, and $192 \times 192$. }
\label{figura::pix2pix_pairs_20}
\end{figure}

\begin{figure}[htb]
\begin{center}
\includegraphics[height=7.5cm,width=15cm]{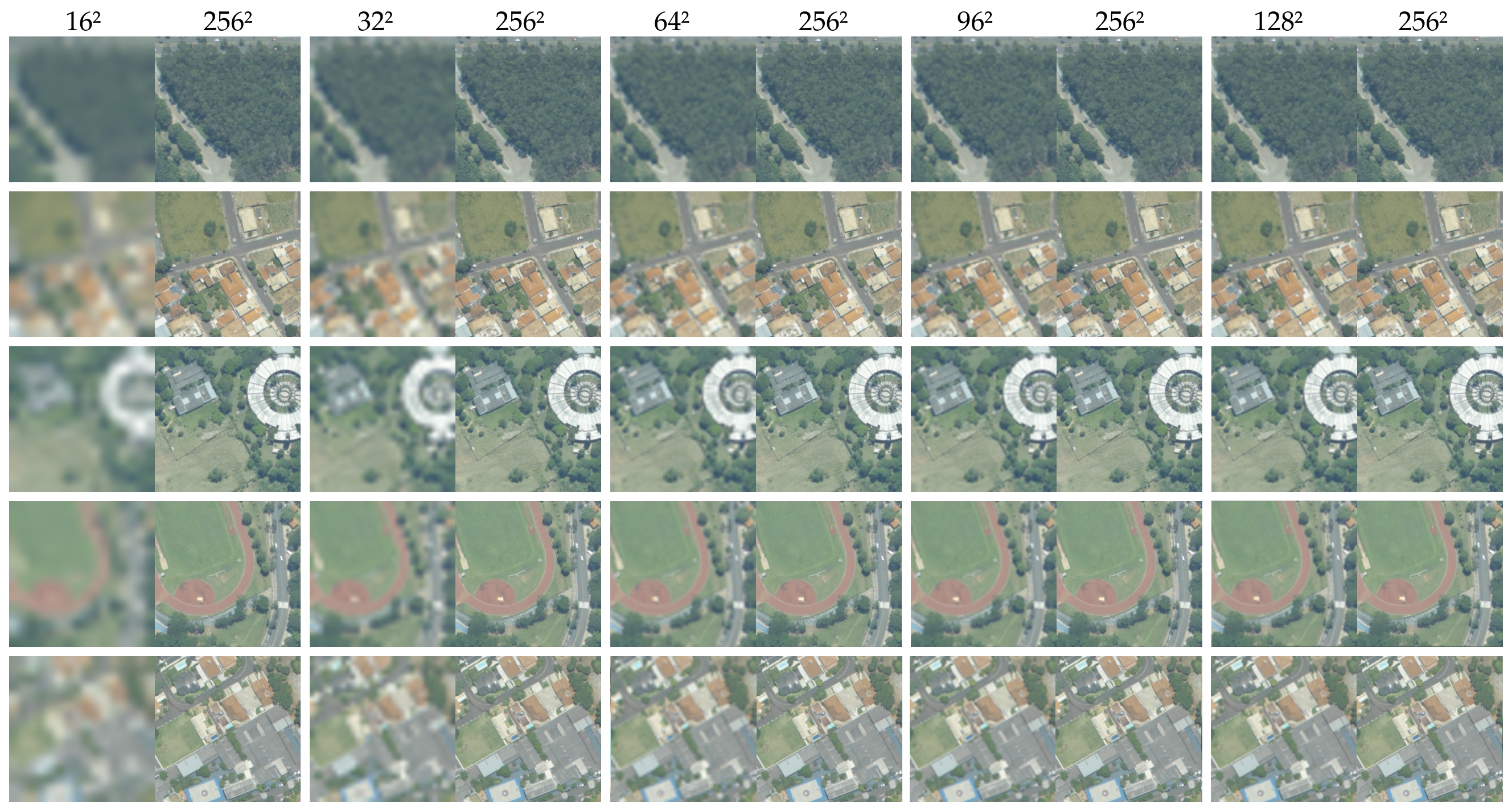}
\end{center}
\caption[Pix2pix training pairs with images of the $P50$ dataset.]{ Pix2pix training pairs with images of the $P50$ dataset at resolutions of $16 \times 16$, $32 \times 32$, $64 \times 64$, $96 \times 96$, and $128 \times 128$. }
\label{figura::pix2pix_pairs_50}
\end{figure}

\begin{figure}[htb]
\begin{center}
\includegraphics[height=15cm,width=15cm]{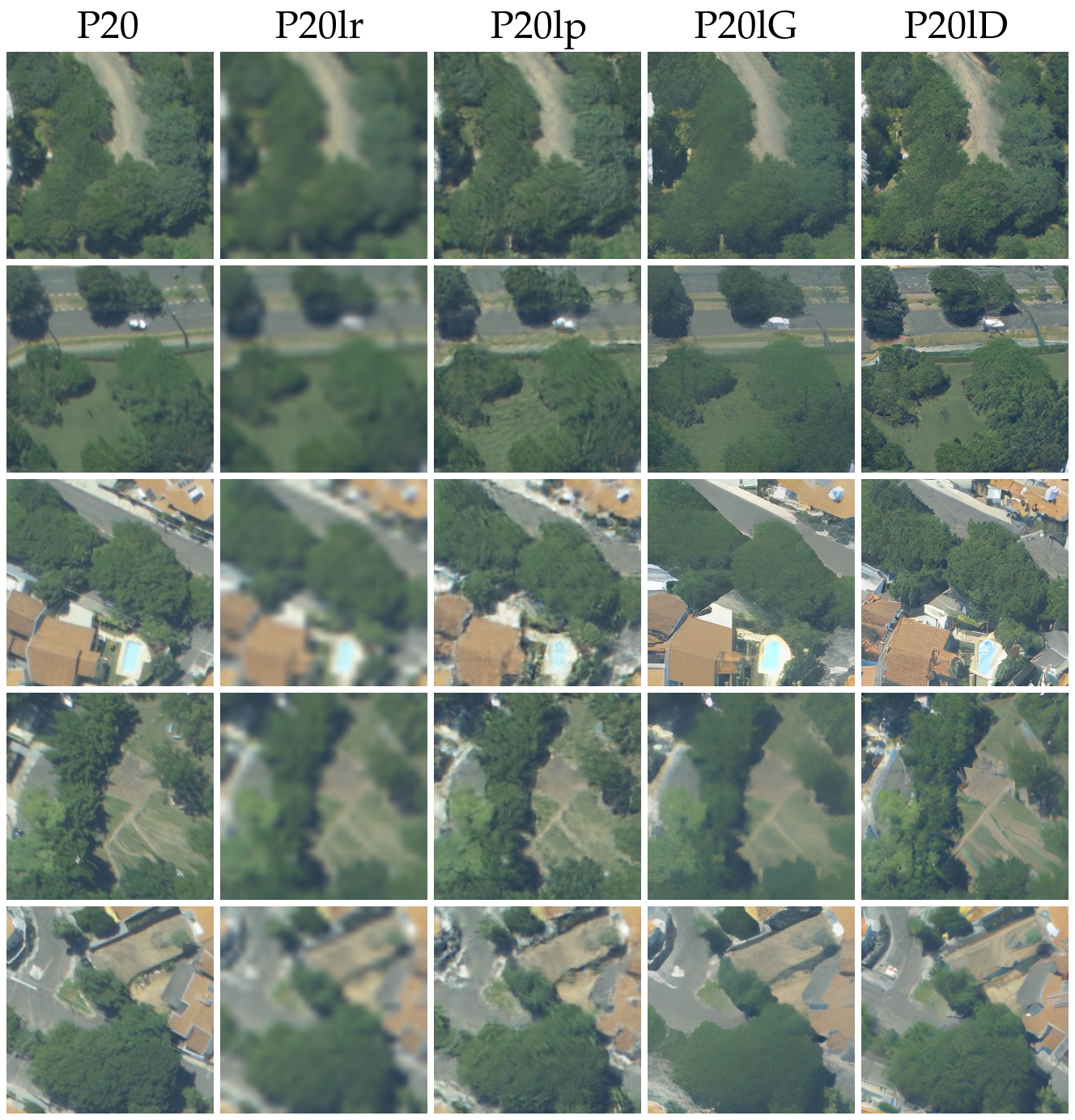}
\end{center}
\caption[Sample images generated using pix2pix, Real-ESRGAN and Diffusion from low resolution images.]{ Sample images generated from low resolution dataset $P20$lr using pix2pix, Real-ESRGAN, and Latent Diffusion}
\label{figura::generated_images_lowres}
\end{figure}

\end{document}